\documentclass[10pt,twocolumn,letterpaper]{article}

\usepackage[pagenumbers]{cvpr} 

\usepackage{amsmath,amsfonts}
\usepackage{array}
\usepackage{textcomp}
\usepackage{stfloats}
\usepackage{url}
\usepackage{verbatim}
\usepackage{graphicx}
\usepackage{cite}
\usepackage{booktabs}
\usepackage{multirow}
\usepackage{soul}
\usepackage{orcidlink}
\usepackage[normalem]{ulem}
\usepackage{color,colortbl}
\usepackage[flushleft]{threeparttable}
\usepackage{balance}
\hypersetup{
    colorlinks=false,
    pdfborder={0 0 0},
}
\definecolor{tree}{rgb}{0.19215686, 0.54509804, 0.34117647}
\definecolor{range}{rgb}{0.0, 1.0, 0.0}
\definecolor{bare}{rgb}{0.50196078, 0.0 , 0.0}
\definecolor{agric1}{rgb}{0.29411765, 0.70980392, 0.28627451}
\definecolor{road1}{rgb}{0.96078431, 0.96078431, 0.96078431}
\definecolor{sealake}{rgb}{0.1372549 , 0.35686275, 0.78431373}
\definecolor{build1}{rgb}{0.96862745, 0.55686275, 0.32156863}

\definecolor{road2}{rgb}{0.65098039, 0.65098039, 0.67058824}
\definecolor{river}{rgb}{0.01176471, 0.02745098, 1.0}
\definecolor{boats}{rgb}{1.0, 0.94901961, 0.0}
\definecolor{agric2}{rgb}{0.66666667, 1.0, 0.0}

\definecolor{vehicle}{rgb}{0.7372549 , 0.83137255, 0.03137255}
\definecolor{parking}{rgb}{0.39215686, 0.4, 0.38823529}
\definecolor{sports}{rgb}{0.36862745, 0.67058824, 0.96862745}
\definecolor{build2}{rgb}{0.65098039, 0.02745098, 0.02745098}

\newcolumntype{C}[1]{>{\centering\let\newline\\\arraybackslash\hspace{0pt}}m{#1}}

\pdfpageheight=\paperheight
\pdfpagewidth=\paperwidth

\title{\vspace{-0.55cm}Generalized Few-Shot Semantic Segmentation in Remote Sensing: \\ Challenge and Benchmark}

\author{Clifford Broni-Bediako$^1$ 
\and Junshi Xia$^1$ 
\and Jian Song$^2$
\and Hongruixuan Chen$^2$
\and Mennatullah Siam$^3$
\and Naoto Yokoya$^{2,1}$ 
\and $^1$RIKEN Center for Advanced Intelligence Project (AIP), Geoinformatics Team, Japan\\
$^2$Department of Complexity Science and Engineering, the University of Tokyo, Japan\\
$^3$Faculty of Engineering and Applied Science, Ontario Tech University, Canada and \\ Computer Science Department, University of British Columbia, Canada
}

\begin{document}
\maketitle
\begin{abstract}
Learning with limited labelled data is a challenging problem in various applications, including remote sensing. Few-shot semantic segmentation is one approach that can encourage deep learning models to learn from few labelled examples for novel classes not seen during the training. The generalized few-shot segmentation setting has an additional challenge which encourages models not only to adapt to the novel classes but also to maintain strong performance on the training base classes. While previous datasets and benchmarks discussed the few-shot segmentation setting in remote sensing, we are the first to propose a generalized few-shot segmentation benchmark for remote sensing. The generalized setting is more realistic and challenging, which necessitates exploring it within the remote sensing context. We release the dataset augmenting OpenEarthMap with additional classes labelled for the generalized few-shot evaluation setting. The dataset is released during the OpenEarthMap land cover mapping generalized few-shot challenge in the L3D-IVU workshop in conjunction with CVPR 2024. In this work, we summarize the dataset and challenge details in addition to providing the benchmark results on the two phases of the challenge for the validation and test sets.
\end{abstract}  
\vspace{-0.15cm}
\section{Introduction}
\label{sec:intro}

Deep learning has shown great success in remote sensing applications with various supervised learning tasks such as land cover mapping (i.e., semantic segmentation)~\cite{demir2018deepglobe,wang2021loveda,toker2022dynamicearthnet}
and crop yield prediction~\cite{you2017deep}. There has also been an emergence of foundation models for remote sensing, which refer to models trained on broad datasets with powerful generalization capabilities~\cite{bommasani2021opportunities}. These remote sensing foundation models focused on either self-supervised learning~\cite{hong2023spectralgpt,stewart2023ssl4eo,cong2022satmae} 
or vision-language modelling~\cite{kuckreja2023geochat,hu2023rsgpt}. On the other hand, few-shot learning which enables deep learning models to learn from few training examples, is still relatively under explored in remote sensing. Although it is of paramount importance specifically with the current release of foundation models and the demonstration of few-shot prompting of such models~\cite{NEURIPS2022_960a172b}. 

Few-shot learning is guided by a few labelled examples (i.e., support set) to generalize to unseen novel classes in the target images (i.e., query set). Most approaches emulate the inference stage during training by sampling pairs of support and query sets. This mechanism is referred to as \textit{meta-learning}. The emergence of foundation models has marked a new paradigm for few-shot learning which explores few-shot prompting of such models~\cite{NEURIPS2022_960a172b}. There have been various works on few-shot learning for the task of semantic segmentation in natural images~\cite{min2021hypercorrelation,siam2020weakly}
where it started with segmenting the novel classes with respect to the background in the query set. A recently proposed {\em generalized few-shot semantic segmentation} setting defines a more realistic scenario where the goal is to perform well on {\em all} classes, novel and base~\cite{tian2022generalized}. This is considerably more challenging than standard few-shot semantic segmentation, yet, to date, there is no dedicated benchmark dataset for generalized few-shot semantic segmentation in remote sensing to the best of our knowledge.
This work explores generalized few-shot semantic segmentation and its intersection with remote sensing, specifically, focusing on submeter-level land cover mapping. We propose a generalized few-shot semantic segmentation benchmark dataset for remote sensing that we release as part of the first challenge of this task, which builds upon the recent dataset, OpenEarthMap~\cite{xia2023openearthmap}. The benchmark dataset and challenge serve to provide a baseline for researchers interested in pursuing learning in low-resource settings for the task of land cover mapping. In summary, our contributions are twofold:
\begin{itemize}
\item We present the OpenEarthMap generalized few-shot semantic segmentation (OEM-GFSS) dataset, a submeter-level land cover mapping dataset, extending the 8 classes of OpenEarthMap~\cite{xia2023openearthmap} to 15 fine-grained classes.
\item We present the first generalized few-shot semantic segmentation benchmark in remote sensing image understanding and provide the baseline in addition to the challenge winners' results.
\end{itemize}

\section{Related Work}
\label{sec:related}

\subsection{Satellite Imagery Datasets}
There has been a plethora of work on remote sensing datasets in deep learning for self-supervised learning~\cite{stewart2023ssl4eo,cong2022satmae}, vision language modelling~\cite{hu2023rsgpt,kuckreja2023geochat} and supervised learning tasks~\cite{johnson2022opensentinelmap,toker2022dynamicearthnet,wang2021loveda,demir2018deepglobe}.
We focus on supervised learning tasks, specifically land cover mapping as a semantic segmentation task. Some of the datasets that are related to our work include OpenSentinalMap~\cite{johnson2022opensentinelmap} and LoveDA~\cite{wang2021loveda} for land cover mapping, DynamicEarthNet~\cite{toker2022dynamicearthnet} for land cover mapping and change detection. Other datasets include DeepGlobe~\cite{demir2018deepglobe} for building, road extraction and land cover mapping. While LoveDA~\cite{wang2021loveda} and DeepGlobe~\cite{demir2018deepglobe} have been adopted for evaluation in cross-domain few-shot semantic segmentation tasks~\cite{Bi2023NotJL,lei2022cross}, yet this work is the first to propose a generalized few-shot semantic segmentation benchmark for remote sensing. Such setting is more realistic as it evaluates both base (i.e., classes used during training) and novel (i.e., classes unseen during training and provided with only few training examples in the few-shot inference).

\subsection{Few-Shot Learning}
Few-shot learning has been heavily investigated in various tasks including classification~\cite{snell2017prototypical,dhillon2019baseline} and segmentation~\cite{siam2020weakly,min2021hypercorrelation}. Few-shot semantic segmentation is focused on purposing few-shot learning for such dense segmentation tasks that require different perspectives than simple classification tasks. For example, few-shot semantic segmentation has seen the prevalence of multiscale approaches~\cite{min2021hypercorrelation} and the exploitation of dense pixel-to-pixel affinities between support and query sets~\cite{siam2020weakly}. While the previous works mainly focused on natural images, there has been some exploration of few-shot learning in remote sensing~\cite{cheng2021spnet,wang2023self,yao2021scale,lei2022cross}. 
Few-shot learning in remote sensing was explored in scene classification~\cite{cheng2021spnet} and in land cover mapping~\cite{wang2023self,yao2021scale,lei2022cross,wang2021loveda}. The most recent few-shot semantic segmentation benchmark on iSAID-5i was released~\cite{yao2021scale} with a focus on segmenting novel classes solely in a 1-way manner, where the models segment the novel class with respect to the background. On the other hand, our benchmark focuses on a more realistic setting for the generalized few-shot semantic segmentation, where models are capable of segmenting both the base and novel classes in an N-way manner. It also provides additional challenges arising from forgetting the base class performance when learning the novel classes from few examples.

\section{Dataset}
\label{sec:dataset}

\begin{figure}
    \centering
    \includegraphics[width=1\linewidth]{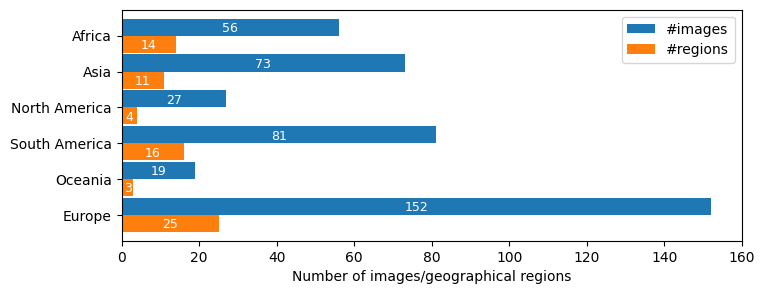}
    \caption{The number of images and geographical regions in the OEM-GFSS dataset across the six continents. There are 408 images from 73 geographical regions. OEM-GFSS has a greater representation in Europe and less in Oceania and North America.}
    \label{fig:regions}
\end{figure}

\begin{figure*}
    \centering
    \includegraphics[width=1\textwidth]{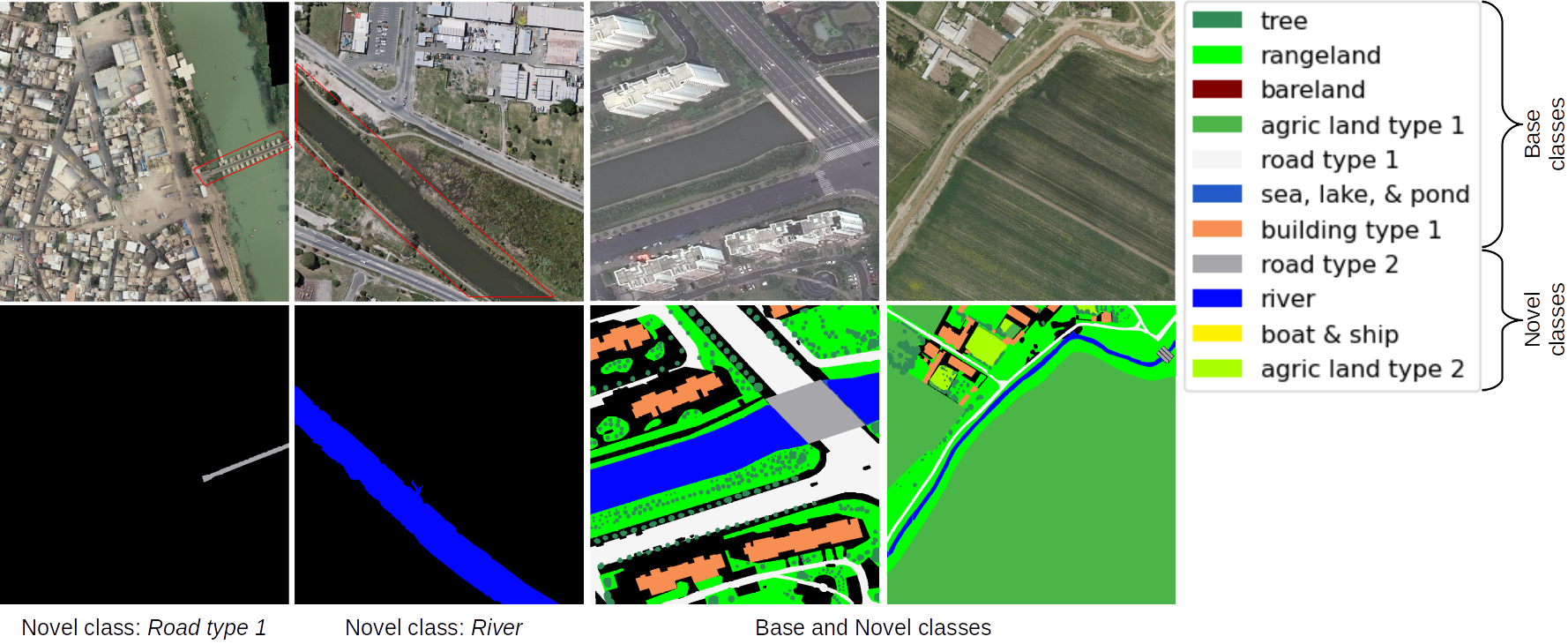}
    \caption{OEM-GFSS validation set examples. The first two columns: examples of novel classes in the support set. The second two columns: base and novel classes in the query set.}
    \label{fig:data-examples}
    \vspace{5mm}
\end{figure*}

This section presents the OpenEarthMap generalised few-shot semantic segmentation (OEM-GFSS) benchmark dataset of remote sensing imagery, which is publicly available\footnote{\url{https://zenodo.org/records/11396874}}.

\subsection{Data Curation and Annotation}\label{sec:3.1}
The OEM-GFSS dataset extends the 8-class coarse-grained land cover labels of the OpenEarthMap dataset~\cite{xia2023openearthmap} to 15-class fined-grained land cover labels. The OEM-GFSS dataset is created from the test set of OpenEarthMap, for which the labels have not been released. We first prepared a set of new classes (see Section~\ref{sec:3.2}) by inspecting all the images, excluding xBD~\cite{Gupta2019CVPR} images, in the test set of OpenEarthMap. Then we sampled images across all the geographical regions in the test set of OpenEarthMap that contain the newly defined classes to create the dataset. This resulted in 408 images that were sampled from 73 regions of the 97 geographical regions across the 6 continents in the OpenEarthMap dataset. Figure~\ref{fig:regions} presents a per-continent image and geographical region counts of the images and regions in OEM-GFSS. The images are of the size of $1024\times1024$ at a spatial resolution of 0.25–0.5m ground sampling distance as in OpenEarthMap. The annotation process follows a similar approach as used in OpenEarthMap, which is manually labelling each pixel of an image by human annotators, and then two additional annotators perform quality checks. If there is a disagreement between the two annotators on a particular labelling, a third person verifies it. All the images were first annotated based on the newly defined classes and the annotations of the OpenEarthMap original class labels were manually modified to yield fine-grained spatial detailed annotations.

\begin{table}
\captionsetup{width=0.98\linewidth}
\caption{Class splits and dataset statistics of the OEM-GFSS dataset.}
\vspace{-3mm}
\label{tab:class-split}
\begin{threeparttable}
\begin{center}
\footnotesize
\setlength\tabcolsep{1pt}
\begin{tabular}{p{1.3cm}p{2.6cm}>{\centering\arraybackslash}p{1.3cm} c c}
    \toprule[.1em]
    \multirow{2}{*}{Data} & \multirow{2}{*}{Class} & \multirow{2}{*}{\#Images} & \#Mask  & Colour \\
    & & & Pixels (K) & (RGB) \\
    \toprule[.1em]
    \multirow{8}{*}{Train set} 
    & Tree & 257 & 43,308 & \colorbox{tree}{\makebox[1.5cm][c]{\textcolor{black}{49, 139, 87}}}\\
    & Rangeland & 254 & 52,174 & \colorbox{range}{\makebox[1.5cm][c]{\textcolor{black}{0, 255, 0}}} \\
    & Bareland & 46 & 2,086 & \colorbox{bare}{\makebox[1.5cm][c]{\textcolor{black}{128, 0, 0}}} \\
    & Agric land type 1 & 133 & 32,711  & \colorbox{agric1}{\makebox[1.5cm][c]{\textcolor{black}{75, 181, 73}}} \\
    & Road type 1 & 248 & 17,654 & \colorbox{road1}{\makebox[1.5cm][c]{\textcolor{black}{245, 245, 245}}} \\
    & Sea, lake, \& pond & 166 & 4,659 & \colorbox{sealake}{\makebox[1.5cm][c]{\textcolor{black}{35, 91, 200}}} \\
    & Building type 1 & 255 & 33,496 & \colorbox{build1}{\makebox[1.5cm][c]{\textcolor{black}{247, 142, 82}}}\\
    \midrule
    \multirow{5}{=}{Validation set} 
    & Road type 2 & 22 & 95  & \colorbox{road2}{\makebox[1.5cm][c]{\textcolor{black}{166, 166, 171}}} \\
    & River & 25 & 1,811 & \colorbox{river}{\makebox[1.5cm][c]{\textcolor{black}{3, 7, 255}}} \\
    & Boat \& ship & 13 & 121 & \colorbox{boats}{\makebox[1.5cm][c]{\textcolor{black}{255, 242, 0}}} \\
    & Agric land type 2 & 17 & 482 & \colorbox{agric2}{\makebox[1.5cm][c]{\textcolor{black}{170, 255, 0}}} \\
    \midrule
    \multirow{5}{*}{Test set} 
    & Vehicle \& cargo-trailer & 85 & 1,005 & \colorbox{vehicle}{\makebox[1.5cm][c]{\textcolor{black}{188, 212, 8}}} \\
    & Parking space & 57 & 1,924  & \colorbox{parking}{\makebox[1.5cm][c]{\textcolor{black}{100, 102, 99}}} \\
    & Sports field & 45 & 1,163 & \colorbox{sports}{\makebox[1.5cm][c]{\textcolor{black}{94, 171, 247}}} \\
    & Building type 2 & 73 & 5,850  & \colorbox{build2}{\makebox[1.5cm][c]{\textcolor{black}{166, 7, 7}}} \\
    \bottomrule[.1em]
    \end{tabular} 
    \begin{tablenotes}
       \item[]{Note: Undefined objects are labelled as \textit{background} with RGB(0, 0, 0).} 
    \end{tablenotes}
    \end{center}
\end{threeparttable}
\end{table}

\subsection{Classes and Data Split}\label{sec:3.2}
\subsubsection{New classes} The annotations were done with the following eight new classes: \textit{vehicle \& cargo-trailer}, \textit{parking space}, \textit{sports field}, \textit{boat \& ship}, \textit{elevated road} (\textit{road type 2}), \textit{non-residential large building} (\textit{building type 2}), \textit{uncultivated agriculture land} (\textit{agric land type 2}), and \textit{sea, lake, \& pond}. The class selection is based on commonly identified land cover objects across the images in the test set of OpenEarthMap.

\subsubsection{OpenEarthMap classes} Three classes were selected without modification: \textit{tree}, \textit{rangeland}, and \textit{bareland}. Based on the newly defined classes, we modified the following four classes: \textit{agriculture land} $\rightarrow$ \textit{cultivated agriculture land} (\textit{agric land type 1}), \textit{road} $\rightarrow$ \textit{non-elevated road} (\textit{road type 1}), \textit{building} $\rightarrow$ \textit{residential \& other building excluding non-residential large building} (\textit{building type 1}), \textit{water} $\rightarrow$ \textit{river}. 

\subsubsection{Dataset splits} We split the 15 classes with a ratio of 7:4:4 for training classes (base), validation novel classes (val-novel), and test novel classes (test-novel), respectively, as disjointed sets. Based on the classes contained in each image, we split the 408 images into 258 as a train set, 50 as a validation set, and 100 as a test set. The train set contains only the images and labels of the base classes and it is for pre-training a backbone network. The validation and test sets contain images and labels of the val-novel and test-novel classes, respectively, and both consist of a \textit{support set} and a \textit{query set} for GFSS task of a 5-shot with 4-novel and 7-base classes. The class splits and dataset statistics are presented in Table~\ref{tab:class-split}, and examples of the OEM-GFSS dataset are shown in Figure~\ref{fig:data-examples}.

\section{Challenge and Benchmark}
\label{sec:benchmark}

In this section, we describe the challenge\footnote{https://cliffbb.github.io/OEM-Fewshot-Challenge/} and the baseline provided to the participants, then we provide the challenge results built on our generalized few-shot segmentation benchmark dataset.

\subsection{Challenge Details}
In order to push the limit on learning with limited labelled data for remote sensing we released our challenge on CodaLab~\footnote{https://codalab.lisn.upsaclay.fr/competitions/17568} that was based on the OpenEarthMap dataset~\cite{xia2023openearthmap}. We hosted our challenge as part of the Learning with Limited Labelled Data for Image and Video Understanding (L3D-IVU) workshop\footnote{https://sites.google.com/view/l3divu2024/overview} in conjunction with the Computer Vision and Pattern Recognition (CVPR) 2024 conference. The challenge was released in two phases. The first phase was the development phase, participants were provided with the training and validation sets, and they were allowed to submit results on the validation set. Additionally, participants had to submit a challenge paper on their proposed method to be eligible to enter the second phase. The second and final phase is the evaluation phase, where participants received the test set and were allowed to submit their results. After evaluation of their final results and based on the novelty of their approach, the top five challenge winners were announced.

\textbf{Baseline:} Our benchmark encompasses the state-of-the-art generalized few-shot segmentation method, DIaM~\cite{hajimiri2023strong}. It is based on a transductive inference mechanism that mainly uses a knowledge distillation term that prevents the base class classifier from forgetting its performance while fine-tuning on the novel classes. We adopt the same setup as state-of-the-art generalized few-shot segmentation methods which we select as our baseline, that operates on natural images~\cite{hajimiri2023strong} with a PSPNet as the architecture used. During base training, the images that contain novel classes are relabelled so that the novel class pixels are set as background, which presents challenges due to ambiguity. During the few-shot inference, novel classes are labelled in the support set in addition to the base classes that were present during training. Due to the nature of remote sensing imagery, each support set image can contain multiple novel classes. We follow a similar procedure to DIaM~\cite{hajimiri2023strong} in the training and the few-shot inference settings. 
For the evaluation, we report mean intersection over union class-wise, mean over the base classes, mean over the novel classes and our final metric which is a weighted sum that gives higher weight to the novel mean. Specifically, for the weighted sum we use the expression $0.4 \times m_{\text{base}} + 0.6 \times m_{\text{novel}}$, where $m_{\text{base}}$ and $m_{\text{novel}}$ are the base and novel mIoUs, respectively. The code for the baseline adapted for the OEM-GFSS challenge is publicly released\footnote{https://github.com/cliffbb/OEM-Fewshot-Challenge}.

\begin{figure}
   \centering
   \includegraphics[width=1\linewidth]{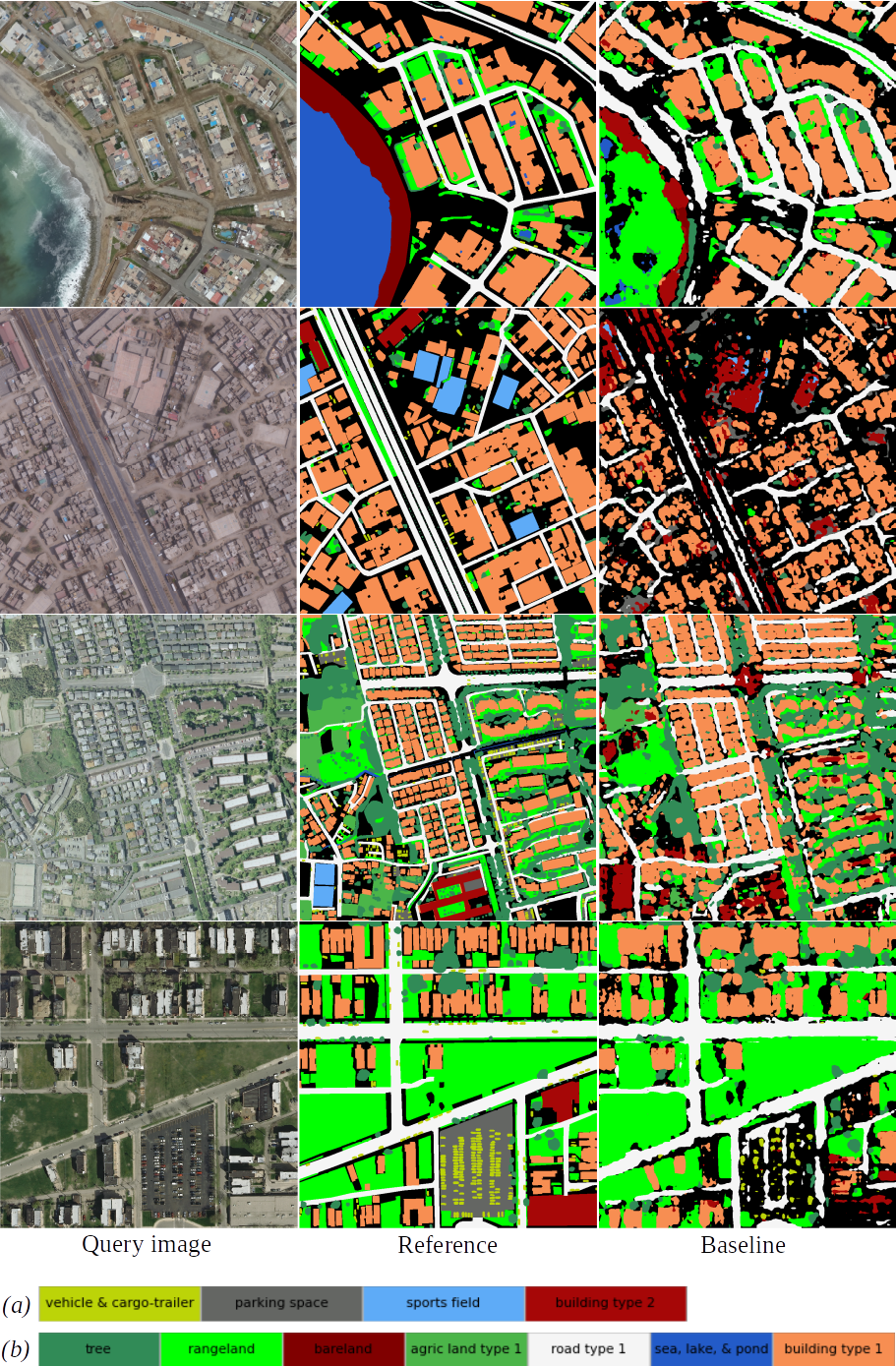}
   \vspace{-3mm}
   \caption{Examples of visual land cover mapping results of the baseline model on the test set of the OEM-GFSS dataset. \textit{(a)} is novel classes of the test set and \textit{(b)} is base classes. Query images can contain both the novel classes and the base classes, and all the classes are to be recognised.}
   \label{fig:baseline-visual-results}
\end{figure}

\begin{table*}
\caption{The Results of the Baseline and the Proposed Methods of the Challenge on the Validation and Test Sets of the OEM-GFSS Dataset. The Boldface indicates Best Results and the Underline indicates the Second Best.}
\vspace{-2.5mm}
\label{tab:challenge_results}
\begin{threeparttable}
\begin{center}
\footnotesize
\scalebox{0.89}{
\setlength\tabcolsep{1pt}
\begin{tabular}{l C{1cm} *{3}{C{1.3cm}} C{1cm}C{1.2cm}*{3}{C{1.4cm}}C{1cm}C{1.4cm}C{0.8cm}C{0.8cm}C{1.3cm}}
    \toprule
    \multicolumn{15}{c}{\textit{Challenge Phase 1 Results (Validation Set)}}\\
    \multirow{4}{*}{Method} & \multicolumn{7}{c}{Base classes (IoU \%)} & \multicolumn{4}{c}{Novel classes (IoU \%)} &  \multirow{4}{*}{\parbox{0.8cm}{\centering Base mIoU (\%)}} & \multirow{4}{*}{\parbox{0.8cm}{\centering Novel mIoU (\%)}} & \multirow{4}{*}{\parbox{1.3cm}{\centering Weighted-Sum mIoU (\%)}}\\ 
    \cmidrule(lr){2-8} \cmidrule(l){9-12}
     & Tree & Rangeland & Bareland & Agric land type 1 & Road type 1 & Sea, lake \& pond & Building type 1 & Road type 2 & River & Boat \& ship & Agric land type 2 &  &  &  \\
    \toprule
     Baseline~\cite{hajimiri2023strong} & 51.48 & 35.15 & 11.57 & 37.78 & 34.68 & 4.80 & 36.86 & 0.20 & 1.56 & 0.00 & 10.24 & 30.33 & 3.00 & 13.93 \\
     SegLand~\cite{Li_2024_CVPR} & \bf{62.69} & \ul{55.52} & \bf{42.79} & \ul{71.56} & \bf{59.29} & 37.84 & 57.56 & \bf{57.06} & \ul{10.86} & 10.76 & 8.22 & \ul{55.33} & 21.72 & 35.17 \\
     ClassTrans~\cite{Wang_2024_CVPR} & \ul{60.19} & \bf{59.96} & \ul{36.69} & \bf{75.82} & \ul{55.21} & 41.78 & \bf{61.45} & 6.94 & 38.13 & 0.00 & \bf{44.78} & \bf{55.88} & 22.46 & \ul{35.83} \\
     FoMA~\cite{Gao_2024_CVPR} & 55.41 & 54.41 & 23.12 & 68.90 & 48.04 & \bf{65.58} & 58.93 & \ul{17.01} & \ul{62.67} & \bf{58.64} & 1.44 & 53.48 & \bf{34.94} & \bf{42.36} \\
     P-SegGPT~\cite{Immanuel_2024_CVPR} & 52.54 & 43.99 & 16.38 & 66.43 & 52.81 & \ul{55.16} & \ul{59.65} & 0.00 & \bf{69.31} & 0.00 & \ul{32.36} & 49.57 & \ul{25.42} & 35.08 \\
     DKA~\cite{Tong_2024_CVPR} & 56.69 & 53.54 & 30.98 & 46.85 & 42.73 & 15.48 & 51.34 & 0.00 & 54.14 & 0.00 & 28.33 & 42.52 & 20.62 & 29.34 \\
     \toprule
    \multicolumn{15}{c}{\textit{Challenge Phase 2 Results (Test Set)}}\\
    \multirow{4}{*}{Method} & \multicolumn{7}{c}{Base classes (IoU \%)} & \multicolumn{4}{c}{Novel classes (IoU \%)} &  \multirow{4}{*}{\parbox{0.8cm}{\centering Base mIoU (\%)}} & \multirow{4}{*}{\parbox{0.8cm}{\centering Novel mIoU (\%)}} & \multirow{4}{*}{\parbox{1.3cm}{\centering Weighted-Sum mIoU (\%)}}\\ 
    \cmidrule(lr){2-8} \cmidrule(l){9-12}
     & Tree & Rangeland & Bareland & Agric land type 1 & Road type 1 & Sea, lake \& pond & Building type 1 & Vehicle \& cargo-trailer & Parking space & Sports field & Building type 2 &  &  &  \\
    \toprule
     Baseline~\cite{hajimiri2023strong} & 58.03 & 32.38 & 0.03 & 38.33 & 37.80 & 0.41 & 41.69 & 11.90 & 0.64 & 0.68 & 23.61 & 29.81 & 9.21 & 17.45 \\
     SegLand~\cite{Li_2024_CVPR} & \bf{69.17} & \bf{53.02} & \ul{30.92} & \bf{62.30} & \bf{63.48} & 53.26 & 61.73 & \bf{45.84} & \bf{49.74} & \bf{55.87} & \bf{61.92} & \bf{56.27} & \bf{53.34} & \bf{54.51} \\
     ClassTrans~\cite{Wang_2024_CVPR} & \ul{68.94} & 49.81 & \bf{32.84} & 53.61 & 57.60 & \ul{53.97} & 55.54 & 37.24 & \ul{32.26} & \ul{49.98} & \ul{52.10} & \ul{53.19} & \ul{42.90} & \ul{47.01} \\
     FoMA~\cite{Gao_2024_CVPR} & 68.39 & 49.14 & 27.72 & \ul{58.92} & 58.09 & \bf{55.86} & \ul{61.81} & \ul{38.01} & 24.76 & 40.59 & 45.60 & 54.28 & 37.24 & 44.05 \\
     P-SegGPT~\cite{Immanuel_2024_CVPR} & 66.69 & \ul{50.69} & 4.65 & 53.94 & \ul{58.66} & 29.06 & \bf{62.40} & 26.39 & 24.65 & 38.50 & 29.73 & 46.58 & 29.82 & 36.52 \\
     DKA~\cite{Tong_2024_CVPR} & 64.51 & 48.32 & 24.00 & 51.70 & 43.24 & 49.69 & 49.86 & 13.04 & 9.09 & 29.19 & 36.78 & 47.33 & 22.02 & 32.15 \\
    \bottomrule
\end{tabular}}
\begin{tablenotes}
   \item[]{Note: The weighted-sum mIoU is calculated using $0.4 \times \textit{base mIoU}  + 0.6 \times \textit{novel mIoU}$.} 
\end{tablenotes}
\end{center}
\end{threeparttable}
\vspace{-4mm}
\end{table*}

\textbf{Challenge winners:}
The first winner of the challenge which we refer to as SegLand~\cite{Li_2024_CVPR}, is based on a precursor generalized few-shot segmentation method to keep the learned prototypes of the novel classes orthogonal to reduce confusion among them while freezing the base class prototypes. They augmented that technique with various strategies including the use of an ensemble of base learners and data augmentation techniques on the few-shot support set, with access to the training set during the few-shot inference. 
The second winner, ClassTrans~\cite{Wang_2024_CVPR}, focused on mining the similarity between base and novel classes to improve the novel class learning, in addition to handling the class imbalance arising. The third winner, FoMA~\cite{Gao_2024_CVPR}, a foundation model assisted framework through multiple strategies to distill, enrich and fuse. The fourth winner, P-SegGPT~\cite{Immanuel_2024_CVPR}, relied on a learnable prompting technique for SegGPT foundation model. Finally, the fifth winner, DKA~\cite{Tong_2024_CVPR}, focused on improving the adaptability to novel classes through efficient parameter tuning and overcoming the catastrophic forgetting on the base classes through relabelling the training set.

\subsection{Challenge Results}
\textbf{Qualitative results:} First, we demonstrate the performance of the baseline on our proposed OEM-GFSS benchmark in Figure~\ref{fig:baseline-visual-results}. It shows four examples with challenging few-shot segmentation tasks, where our baseline is struggling to segment the novel classes such as parking space and vehicle/cargo trailer. Yet, the baseline is performing relatively well in segmenting the base classes except in certain instances such as the first example sea/lake/pond which was confused for a range land. Nonetheless, the baseline demonstrated relatively well performance overall both on the novel and base class performance, due to its transductive inference that was able to cope with the challenges presented in the test set.

\textbf{Quantitative results:}
Table~\ref{tab:challenge_results} shows the challenge results for the two phases, where the first was evaluated on the validation set and the second evaluated on the test set. It shows the IoU per class for both the base and novel classes, the mean IoU for the base and novel, and the weighted average which was used as the final score to rank the winning entries. In the first phase, it shows that FoMA was the winning entry, where it outperformed all the other methods in the mIoU of the novel classes and the final weighted average. FoMA relied on enriching labels and distilling knowledge from a vision language foundation model which resulted in such performance. It also shows ClassTrans outperforming all other methods in the base classes IoU due to handling the class imbalance in the dataset. In the second phase, SegLand outperformed all the other methods in the ranking score with a considerable margin due to the ensemble of learners.

\section{Conclusion}
We presented our challenge and benchmark for generalized few-shot semantic segmentation in remote sensing, OEM-GFSS, towards encouraging models adaptability to novel classes beyond the closed set of training classes. Our challenge had five winning entries, where we presented their results in both phases in addition to the baseline quantitative and qualitative results. By making our benchmark publicly available, it will foster more research on the challenging problem of learning with limited labelled data in the context of remote sensing.

{
    \small
    \bibliographystyle{ieeenat_fullname}
    \bibliography{main}
}

\end{document}